\documentclass[review]{elsarticle}

\usepackage{times}
\usepackage{helvet}
\usepackage{courier}
\usepackage{graphicx}
\usepackage{booktabs}
\usepackage{amsmath}
\usepackage{amssymb}
\usepackage{subfigure}
\usepackage{algorithm}
\usepackage{algorithmic}

\def\0{{\bf 0}}
\def\1{{\bf 1}}

\usepackage{hyperref}

\journal{Journal of Finance and Data Science}

\begin{document}

\begin{frontmatter}

\title{An Overview on Data Representation Learning: From Traditional Feature Learning to Recent Deep Learning}

\author{Guoqiang Zhong\corref{mycorrespondingauthor}, Li-Na Wang, Junyu Dong}


\address{Department of Computer Science and Technology,\\
  Ocean University of China, \\
  238 Songling Road, Qingdao 266100, China \\
  Email: gqzhong@ouc.edu.cn}

\begin{abstract}
Since about 100 years ago, to learn the intrinsic structure of data, many representation learning approaches have been proposed, including both linear ones and nonlinear ones, supervised ones and unsupervised ones. Particularly, deep architectures are widely applied for representation learning in recent years, and have delivered top results in many tasks, such as image classification, object detection and speech recognition. In this paper, we review the development of data representation learning methods. Specifically, we investigate both traditional feature learning algorithms and state-of-the-art deep learning models. The history of data representation learning is introduced, while available resources (e.g. online course, tutorial and book information) and toolboxes are provided. Finally, we conclude this paper with remarks and some interesting research directions on data representation learning.
\end{abstract}


\end{frontmatter}


\section{Introduction}

In many domains, such as artificial intelligence, bioinformatics and finance, data representation learning is a critical step to facilitate the subsequent classification, retrieval and recommendation tasks. Typically, for large scale applications, how to learn the intrinsic structure of data and discover valuable information from data becomes more and more important and challenging.

Since about 100 years ago, many data representation learning methods have been proposed. Among others, principal component analysis (PCA) was proposed by K. Pearson in 1901~\cite{PCA}, while linear discriminant analysis (LDA) was proposed by R. Fisher in 1936~\cite{LDA}. PCA and LDA are both linear methods. Nevertheless, PCA is an unsupervised method, whilst LDA is a supervised one. Based on PCA and LDA, variety of extensions have been proposed, such as kernel PCA~\cite{ScholkopfSM98} and generalized discriminant analysis (GDA)~\cite{GDA}. In 2000, the machine learning community launched the research on manifold learning, which is to discover the intrinsic structure of high dimensional data. Unlike previous global approaches, such as PCA and LDA, manifold learning methods are generally locality based, such as isometric feature mapping (Isomap)~\cite{ISOMAP} and locally linear embedding (LLE)~\cite{LLE}. In 2006, G. Hinton and his co-authors successfully applied deep neural networks to dimensionality reduction, and proposed the concept of ``deep learning"~\cite{HintonOT06,Hinton504}. Nowadays, due to their high effectiveness, deep learning algorithms have been employed in many areas beyond artificial intelligence.

On the other hand, the research on artificial neural networks undergoes a tough process, with many successes and difficulties. In 1943, W. McCulloch and W. Pitts created the first artificial neuron, linear threshold unit, which is also called M-P model in the following research~\cite{MPmodel}, for neural networks. Later, D. Hebb proposed a hypothesis of learning based on the mechanism of neural plasticity, which is also known as Hebbian theory~\cite{Hebb}. Essentially, M-P model and Hebbian theory paved the way for neural network research and the development of connectionism in the area of artificial intelligence. In 1958, F. Rosenblatt created the perceptron, a two-layer neural network for binary classification~\cite{Perceptron}. However, M. Minsky and S. Papert pointed out that perceptrons were even incapable of solving the exclusive-or (XOR) problem~\cite{Minsky}. Until 1974, P. Werbos proposed the back propagation algorithm to train multi-layer perceptrons (MLP)~\cite{Werbos}, the neural network research had stagnated. Particularly, in 1986, D. Rumelhart, G. Hinton and R. Williams showed that the back propagation algorithm can generate useful internal representations of data in hidden layers of neural networks~\cite{Rumelhart}. With the back propagation algorithm, although one could train many layers of neural networks in theory, two crucial issues existed: model overfitting and gradient diffusion. In 2006, G. Hinton initiated the breakthrough on representation learning research with the idea of greedy layer-wise pre-training plus finetuing of deep neural networks~\cite{HintonOT06,Hinton504}. The issues confusing the neural network community were addressed accordingly. Later on, many deep learning algorithms were proposed and successfully applied to various domains~\cite{BengioLPL06,RanzatoPCL06}.

In this paper, we review the development of data representation learning, i.e. both traditional feature learning and recent deep learning. The rest of this paper is organized as follows. Section 2 is devoted to traditional feature learning, including linear algorithms and their kernel extension, and manifold learning methods. In Section 3, we introduce the recent progress of deep learning, including important models and public toolboxes. Section 4 concludes this paper with remarks and interesting research directions on data representation learning.

\section{Traditional feature learning}

In this section, we focus on traditional feature learning algorithms, which belong to ``shallow" models and aim to learn transformations of data that make it easier to extract useful information when building classifiers or other predictors~\cite{BengioCV13}. Hence, we will not consider some manual feature engineering methods, such as image descriptors (e.g. scale-invariant feature transform or SIFT~\cite{Lowe99}, local binary patterns or LBP~\cite{OjalaPH96}, and histogram of oriented gradients or HOG~\cite{DalalT05}, and so on) and document statistics (e.g. term frequency-inverse document frequency or TF-IDF~\cite{tfidf}, and so on).

From the perspective of its formulation, an algorithm is generally considered to be linear or nonlinear, supervised or unsupervised, generative or discriminative, global or local. For example, PCA is a linear, unsupervised, generative and global feature learning method, while LDA is a linear, supervised, discriminative and global method. In this section, we adopt the taxonomy to categorize the feature learning algorithms as global ones or local ones. In general, global methods try to preserve the global information of data in the learned feature space, but local ones focus on preserving local similarity between data during learning the new representations. For instance, unlike PCA and LDA, LLE is a locality-based feature learning algorithm. Moreover, we usually call locality-based feature learning as manifold learning, since it is to discover the manifold structure hidden in the high dimensional data.

In the literature, van der Maaten, Postma and van den Herik provided a MATLAB toolbox for dimensionality reduction, which includes the codes of 34 feature learning algorithms~\cite{drtoolbox}. In~\cite{GE}, Yan et al. introduced a general framework known as graph embedding to unify a large family of dimensionality reduction algorithms into one formulation. In~\cite{ZhongCC13}, Zhong, Chherawala and Cheriet compared three kinds of supervised dimensionality reduction methods for handwriting recognition, while in~\cite{ZhongC15}, Zhong and Cheriet presented a framework from the viewpoint of tensor representation learning, which considers the input data as tensors and unifies many linear, kernel and tensor dimensionality reduction methods with one learning criterion.

\subsection{Global feature learning}

As mentioned above, PCA is one of the earliest linear feature learning algorithm~\cite{PCA,Joliffepca}. Due to its simplicity, PCA has been widely used for dimensionality reduction~\cite{eigenface}. It uses an orthogonal transformation to convert a set of observations of possibly correlated variables into a set of values of linearly uncorrelated variables. To some extent, classical multidimensional scaling (MDS) is similar with PCA, i.e. both of them are linear method and optimized using eigenvalue decomposition~\cite{MDS}. The difference between PCA and MDS is that, the input of PCA is the data matrix, while that of MDS is the distance matrix between data. Except for eigenvalue decomposition, singular value decomposition (SVD) is often used for optimization as well. Latent semantic analysis (LSA) in information retrieval is just optimized using SVD, which reduces the number of rows while preserving the similarity structure among columns (rows represent words and columns represent documents)~\cite{Dumais04}. As variants of PCA, kernel PCA extends PCA for nonlinear dimensionality reduction using the kernel trick~\cite{KPCA}, while probabilistic PCA is a probabilistic version of PCA~\cite{PPCA}. Moreover, based on PPCA, Lawrence proposed the Gaussian process latent variable model (GPLVM), which is a fully probabilistic, nonlinear latent variable model and can learn a nonlinear mapping from the latent space to the observation space~\cite{Lawrence05}. In order to integrate supervisory information into the GPLVM framework, Urtasun and Darrell proposed the discriminative GPLVM~\cite{DGPLVM}. However, since DGPLVM is based on the learning criterion of LDA~\cite{LDA} or GDA~\cite{GDA}, the dimensionality of the learned latent space in DGPLVM is restricted to be at most $C - 1$, where $C$ is the number of classes. To address this problem, Zhong et al. proposed the Gaussian process latent random field (GPLRF)~\cite{ZhongLYHL10}, by enforcing the latent variables to be a Gaussian Markov random field (GMRF) \cite{GMRFbook} with respect to a graph constructed from the supervisory information. Among others, some more extensions of PCA include sparse PCA~\cite{spca}, robust PCA~\cite{CandesLMW11,BouwmansZ14} and probabilistic relational PCA~\cite{LiYZ09}.

LDA is a supervised, linear feature learning method, which enforces data belonging to the same class to be close and that belonging to different classes to be far away in the learned low-dimensional subspace~\cite{LDA}. LDA has been successfully used in face recognition, and the obtained new features are called Fisherfaces~\cite{BelhumeurHK97}. GDA is the kernel version of LDA~\cite{GDA}. In general, LDA and GDA are learned with the generalized eigenvalue decomposition. However, Wang et al. pointed out that the solution of generalized eigenvalue decomposition was only an approximation to that of the original trace ratio problem with respect to LDA's formulation~\cite{WangCVPR}. Hence, they transformed the trace ratio problem to a series of trace difference problems and used an iterative algorithm to solve it. Furthermore, Jia et al. put forward a novel Newton-Raphson method for trace ratio problems, which can be proved to be convergent~\cite{Jiatr}. In~\cite{ZhongSC16}, Zhong, Shi and Cheriet have presented a novel method called relational Fisher analysis, which is based on the trace ratio formulation and sufficiently exploits the relational information of data. In~\cite{ZhongL16}, Zhong and Ling analyzed an iterative algorithm for the trace ratio problems and proved the necessary and sufficient conditions for the existence of the optimal solution of trace ratio problems. In addition, more extensions of LDA may include incremental LDA~\cite{GhassabehRM15}, DGPLVM~\cite{DGPLVM} and marginal Fisher analysis (MFA)~\cite{GE}.

Except feature learning algorithms mentioned above, there are many other feature learning methods, such as independent component analysis (ICA)~\cite{ICA}, canonical-correlation analysis (CCA)~\cite{HOTELLING01121936}, ensemble learning based feature extraction~\cite{ZhongL13}, multi-task feature learning~\cite{AECOC}, and so on. Moreover, to directly process tensor data, many tensor representation learning algorithms have been proposed~\cite{YangZFY04, YeJL04, HeCN05a, GE, FuH08, ZhongC12, ZhongC14, JiaZF14, JiaKDF14, Zhong2014}. For example, Yang et al. proposed the 2DPCA algorithm and shew its advantage over PCA on face recognition problems~\cite{YangZFY04}, while Ye, Janardan and Li proposed the 2DLDA algorithm, which extends LDA for two-order tensor representation learning~\cite{YeJL04}. Particularly, in~\cite{ZhongC14}, a large margin low rank tensor representation learning algorithm is introduced, the convergence of which can be theoretically guaranteed.

\subsection{Manifold learning}

In this section, we focus on locality-based feature learning methods, and call them manifold learning methods. Although most of the manifold learning algorithms are nonlinear dimensionality reduction approaches, some are linear dimensionality reduction methods, such as locality preserving projections~\cite{LPP} and MFA~\cite{GE}. Meanwhile, note that some nonlinear dimensionality reduction algorithms are not manifold learning approaches, as they are not aimed to discover the intrinsic structure of high dimensional data, such as Sammon mapping~\cite{Sammon} and KPCA~\cite{KPCA}.

In 2000, ``\emph{Science}" published two interesting papers on manifold learning. The first paper introduces Isomap, which combines the Floyd-Warshall algorithm with classic MDS~\cite{ISOMAP}. Based on local neighborhood of the samples, Isomap computes the pair-wise distance between data using the Floyd-Warshall algorithm, and then, learn the low-dimensional embeddings of data using classic MDS on the computed pair-wise distances. The second paper is about LLE, which encodes the locality information at each point into the reconstruction weights of its neighbors. Later on, many manifold learning algorithms were proposed~\cite{LE, Donoho, ZhangZ04, Diffusion, WeinbergerS06, LPP, GE, WangM08}. In particular, the work of~\cite{LTSLE} combines the idea of local tangent space alignment (LTSA)~\cite{ZhangZ04} and Laplacian eigenmaps (LE)~\cite{LE}, which computes the local similarity between data using the Euclidean distance in the local tangent space and employs LE to learn the low dimensional embeddings of data. In~\cite{MohamedChap}, Cheriet et al. applied manifold learning approaches to shape-based recognition of historical Arabic documents and obtained noticeable improvement over previous methods.

In addition to the methods mentioned above, some related work that needs pay attention to includes the algorithms for distance metric learning~\cite{EricXing, WeinbergerBS05, ZhongHL11, ZhongZLF16}, semi-supervised learning~\cite{BelkinNS06}, dictionary learning~\cite{LeeBRN06}, and non-negative matrix factorization~\cite{DhillonS05}, which to some extent take account of the underlying structure of data.

\section{Deep learning}

In the literature, 4 survey papers on deep learning have been published. In~\cite{Bengio09}, Bengio introduced the motivations, principles and some important algorithms of deep learning, while in~\cite{BengioCV13}, from the perspective of representation learning, Bengio, Courville and Vincent reviewed the progress of feature learning and deep learning. In~\cite{DLNature}, LeCun, Bengio and Hinton introduced the development of deep learning and some important deep learning models including convolutional neural network~\cite{Convnet} and recurrent neural network~\cite{RNN}. In~\cite{Schmidhuber15}, Schmidhuber reviewed the development of the artificial neural networks and deep learning year by year. With these survey papers, the readers who are interested to deep learning may easily understand the research area of deep learning and its history.

To learn deep learning algorithms, some internet resources are worth being recommended. The first one is the Coursera course taught by Professor Hinton. Its webpage is at https://www.coursera.org/learn/neural-networks\#. This course is about artificial neural networks and how they're being used for machine learning. The second one is the tutorial on unsupervised feature learning and deep learning, provided by some researchers at Stanford University. Its webpage is at http://ufldl.stanford.edu/wiki/index.php/UFLDL\_Tutorial. Except basic knowledge on unsupervised feature learning and deep learning algorithms, this tutorial includes many exercises. Hence, it's quite suitable for deep learning beginners. The third one is the deep learning website. It's at http://deeplearning.net/. This website provides not only deep learning tutorials, but also reading list, softwares, data sets and so on. The fourth one is a blog, which is written in Chinese. It's at http://www.cnblogs.com/tornadomeet/. The host of this blog records the process how she/he learned deep learning and wrote the codes model by model. Nevertheless, many other blogs and webpages are also useful and helpful, such as http://blog.csdn.net/ and Wikipedia. The last but not the least is the deep learning book written by Professor Goodfellow, Bengio and Courville, which has been published by MIT Press. Its online version is free and the webpage is at http://www.deeplearningbook.org/. With these courses, tutorials, blogs and books, the students and engineers who may study or work on deep learning can basically understand the theoretical details of the deep learning algorithms.

\subsection{Deep learning models}

Here, we review some deep learning models, especially that proposed after the publication of~\cite{BengioCV13}.

The renewal of deep learning is mainly due to the great progress of three aspects: feature learning, availability of large scale of labeled data, and hardware, especially general purpose graphics processing units (GPGPU). In 2006, Hinton and his colleagues proposed to use greedy layer-wise pre-training and finetuning for the learning of deep neural networks, which results in higher performance than state-of-the-art algorithms on MNIST handwritten digits recognition and document retrieval tasks~\cite{Hinton504}. Quickly, some work followed up. Bengio et al. introduced the stacked autoencoders and confirmed the hypothesis that the greedy layer-wise unsupervised training strategy mostly helps the optimization, by initializing weights in a region near a good local minimum, giving rise to internal distributed representations that are high-level abstractions of the input, and bringing better generalization~\cite{BengioLPL06}. In~\cite{VincentLLBM10}, Vincent et al. proposed the stacked denoising autoencoders, which are trained locally to denoise corrupted versions of the inputs. In~\cite{ZhengZLCD14}, Zheng et al. showed the effectiveness of deep architectures that are built with stacked feature learning modules, such as PCA and stochastic neighbor embedding (SNE)~\cite{HintonR02}. To improve the effectiveness of the deep architectures built with stacked feature learning models, Zheng et al. applied the stretching technique~\cite{PandeyD14} on the weight matrix between the top successive layers, and demonstrated the effectiveness of the proposed method on handwritten text recognition tasks~\cite{ZhengCZCSD15}. Additionally, in~\cite{Partha}, a tandem hidden Markov model using deep belief networks (DBNs)~\cite{HintonOT06} was proposed and applied for offline handwriting recognition.

In 2012, Krizhevsky, Sutskever and Hinton created the ``AlexNet" and won the ImageNet Large Scale Visual Recognition Competition (ImageNet LSVRC)~\cite{KrizhevskySH12}. In AlexNet, the dropout regularization~\cite{abs-1207-0580} and the nonlinear activation function called rectified linear units (ReLUs)~\cite{NairH10} were used. To speed up the learning on 1.2 million training images from 1000 categories, AlexNet was implemented on GPUs. Between 2013 and 2016, all the models performed well in the ImageNet LSVRC are based on deep convolutional neural networks (CNNs), such as OverFeat~\cite{SermanetEZMFL13}, VGGNet~\cite{SimonyanZ14a}, GoogLeNet~\cite{SzegedyLJSRAEVR14} and ResNet~\cite{HeZRS15}. In~\cite{DonahueJVHZTD14}, an interesting feature extraction method based on AlexNet was proposed. The authors showed that features extracted from the activation of a deep convolutional network (e.g. AlexNet) trained in a fully supervised fashion on a large, fixed set of object recognition tasks can be repurposed to novel generic tasks. Accordingly, this feature was called deep convolutional activation feature (DeCAF). In~\cite{classdocimgs}, Zhong et al. introduced two challenging problems related to photographed document images and applied DeCAF to set the baseline results for the proposed problems. In~\cite{CaiZZHD15}, Cai et al. considered the problem that whether DeCAF is good enough for accurate image classification, and based on the reducing and stretching operations, the authors improved DeCAF on several image classification problems. Based on the AlexNet and VGGNet, Zhong et al. proposed a deep hashing learning algorithm, which greatly improved previous hashing learning approaches on image retrieval.

Recently, the deep learning models obtained much attention are recurrent neural networks (RNNs)~\cite{RNN, GravesLFBBS09}, long short-term memory (LSTM)~\cite{HochreiterS97, WandKS16}, attention based models~\cite{ChorowskiBSCB15, BahdanauCSBB16} and generative adversarial nets~\cite{GoodfellowPMXWOCB14}. The applications are generally focused on image classification, object detection, speech recognition, handwriting recognition, image caption generation and machine translation~\cite{DengL13, XuBKCCSZB15, SutskeverVL14}.

\subsection{Deep learning toolboxes}

There are many deep learning toolboxes commonly shared on the internet. In each toolbox, the codes of some deep learning models, such as DBNs~\cite{HintonOT06}, LeNet-5~\cite{Convnet}, AlexNet~\cite{KrizhevskySH12} and VGGNet~\cite{SimonyanZ14a}, are often provided, respectively. The researchers may directly use the codes or develop new models based on the codes under certain licenses. In the following, we briefly introduce Theano\footnote{http://www.deeplearning.net/software/theano/index.html}, Caffe\footnote{http://caffe.berkeleyvision.org/}, TensorFlow\footnote{https://www.tensorflow.org/} and MXNet\footnote{http://mxnet.io/index.html}.

Theano is a Python library. It is tightly integrated with NumPy, and allows the users to define, optimize, and evaluate mathematical expressions involving multi-dimensional arrays efficiently. Moreover, it could perform data-intensive calculations on GPUs with up to 140 times faster than with CPU. The deep learning tutorial provided at http://deeplearning.net/tutorial/ is just based on Theano. Caffe is a pure C++/CUDA toolbox for deep learning. However, it provides command line, Python and MATLAB interfaces. The Caffe codes run fast, and can seamless switch between CPU and GPU. TensorFlow is an open source software library for numerical computation using data flow graphs. Nodes in the graph represent mathematical operations, while the graph edges represent the multidimensional data arrays (tensors) communicated between them. TensorFlow has the automatic differentiation capability to facilitate the computation of derivatives. MXNet is developed by many collaborators from several universities and companies. It supports both imperative and symbolic programming, and multiple programming languages, such as C++, Python, R, Scala, Julia, Matlab and Javascript. In general, the running speed of MXNet codes is comparative with that of Caffe codes, and much faster than that of Theano and TensorFlow.

\section{Conclusion}

In this paper, we review the research on data representation learning, including traditional feature learning and recent deep learning. From the development of feature learning methods and artificial neural networks, we can see that deep learning is not totally new. It's the consequence of the great progress of feature learning research, availability of large scale of labeled data, and hardware. However, the breakthrough of deep learning not only affects the artificial intelligence area, but also greatly improves the progress of many domains, such as finance~\cite{HeatonPW16} and bioinformatics~\cite{MinLY16}.

For the future research on deep learning, we provide three directions: the fundamental theory, novel algorithms, and applications. Some researchers have tried to analyze deep neural networks~\cite{ErhanBCMVB10, CohenSS16, EldanS16}. However, the gap between the theory and application of deep learning is still quite large. Although many deep learning algorithms have been proposed, most of them are based on deep CNNs or RNNs. Therefore, creative deep learning algorithms need to be proposed, to solve real world problems, such as unsupervised models and transfer learning models. Moreover, deep learning algorithms have been preliminarily exploited in many domains. However, to solve some challenging problems, such as that in natural language processing and computer vision, more sophisticated models need to be created and applied.

Finally, we emphasize that deep learning is not the whole of machine learning, and the only way to realize artificial intelligence. To solve the real world problems, many approaches for intelligent data analytics are necessary.

\section*{Acknowledgment}\label{sec:acknowledge}

This work was supported by the National Natural Science Foundation of China (NSFC) under
Grant No. 61403353, the Open Project Program of the National Laboratory of Pattern
Recognition (NLPR) and the Fundamental Research Funds for the Central Universities of
China.

\section*{References}\label{sec:refs}

\bibliographystyle{model1-num-names}
\bibliography{mybibfile}

\end{document}